\def\year{2022}\relax
\documentclass[letterpaper]{article} 
\usepackage{aaai22}  
\usepackage{times}  
\usepackage{helvet}  
\usepackage{courier}  
\usepackage[hyphens]{url}  
\usepackage{graphicx} 
\usepackage{natbib}  
\usepackage{caption} 
\DeclareCaptionStyle{ruled}{labelfont=normalfont,labelsep=colon,strut=off} 
\usepackage{amsmath}
\usepackage{bm}
\usepackage{algorithm}
\usepackage{algorithmic}

\usepackage{newfloat}
\usepackage{listings}
\floatstyle{ruled}
\newfloat{listing}{tb}{lst}{}
\floatname{listing}{Listing}
\usepackage{subfig}
\usepackage{amssymb}
\usepackage{booktabs}

\usepackage{multirow}
\title{Inharmonious Region Localization by Magnifying Domain Discrepancy}
\author{
    Jing Liang\textsuperscript{\rm 1}, Li Niu\textsuperscript{\rm 1}\thanks{Corresponding Author.}, Penghao Wu\textsuperscript{\rm 1}, Fengjun Guo\textsuperscript{\rm 2}, Teng Long\textsuperscript{\rm 2} 
}
\affiliations{
    \textsuperscript{\rm 1}MoE Key Lab of Artificial Intelligence, Shanghai Jiao Tong University\\
   \textsuperscript{\rm 2}INTSIG \\
\{leungjing,ustcnewly,wupenghaoCraig\}@sjtu.edu.cn, \{fengjun\_guo, mike\_long\}@intsig.net
}

\usepackage{bibentry}

\begin{document}
\maketitle

\begin{abstract}
Inharmonious region localization aims to localize the region in a synthetic image which is incompatible with surrounding background. The inharmony issue is mainly attributed to the color and illumination inconsistency produced by image editing techniques. In this work, we tend to transform the input image to another color space to magnify the domain discrepancy between inharmonious region and background, so that the model can identify the inharmonious region more easily. To this end, we present a novel framework consisting of a color mapping module and an inharmonious region localization network, in which the former is equipped with a novel domain discrepancy magnification loss and the latter could be an arbitrary localization network. Extensive experiments on image harmonization dataset show the superiority of our designed framework. Our code is available at \url{https://github.com/bcmi/MadisNet-Inharmonious-Region-Localization}.
\end{abstract}

\section{Introduction}

With the rapid development of image editing techniques and tools (\emph{e.g.}, appearance adjustment, copy-paste), users can blend and edit existing source images to create fantastic images that are only limited by an artist's imagination. 
However, some manipulated regions in the created synthetic
images may have inconsistent color and lighting statistics
with the background, which could be attributed to careless editing or the difference among source images (\emph{e.g.}, capture condition, camera setting, artistic style). We refer to such regions as inharmonious regions~\cite{liang2021inharmonious}, which will remarkably downgrade the quality and fidelity of synthetic images.

Recently, the task of inharmonious region localization~\cite{liang2021inharmonious} has been proposed to identify the inharmonious regions. When the inharmonious regions are identified, users can manually adjust the inharmonious regions or employ image harmonization methods~\cite{tsai2017deep,cong2020dovenet,cun2020improving,cong2021bargainnet} to harmonize the inharmonious regions, yielding the images with higher quality and fidelity.

To the best of our knowledge, the only existing inharmonious region localization method is DIRL~\cite{liang2021inharmonious}, which attempted to fuse multi-scale features and avoid redundant information. However, DIRL is a rather general model without exploiting the uniqueness of this task, that is, the discrepancy between inharmonious region and background. Besides, the performance of DIRL is still far from satisfactory when the inharmonious region is surrounded by cluttered background or objects that have similar shapes to the inharmonious region. 

\begin{figure}[t]
\centering
\includegraphics[width=0.9\linewidth]{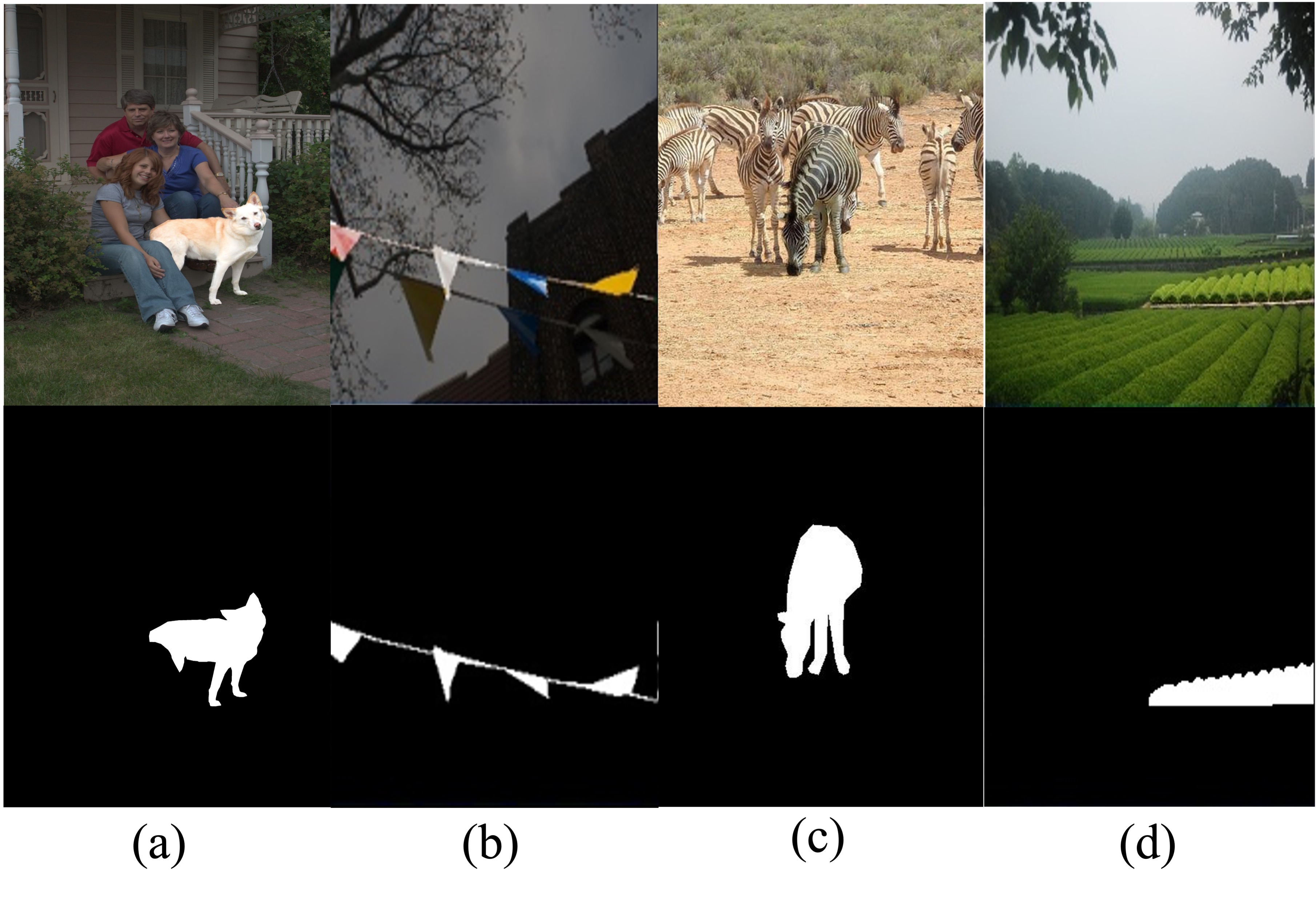} 
\caption{We show the examples of inharmonious synthetic images in the top row and their inharmonious region masks in the bottom row.} 
\label{fig:teaser}
\end{figure}

Considering the uniqueness of inharmonious region localization task, we refer to each suite of color and illumination statistics as one domain following~\cite{cong2020dovenet,cong2021bargainnet}. Thus, the inharmonious region and the background belong to two different domains. In this work, we propose a novel method based on a simple intuition: \emph{can we transform the input image to another color space to magnify the domain discrepancy between inharmonious region and background, so that the model can identify the inharmonious region more easily? }

To achieve this goal, we propose a framework composed of two components: one color mapping module and one inharmonious region localization network.
First, the color mapping module transforms the input image to another color space. Then, the inharmonious region localization network detects the inharmonious region based on the transformed image.
For color mapping module, we extend  HDRNet~\cite{gharbi2017deep} to improved HDRNet (iHDRNet). HDRNet is popular and has achieved great success in previous works~\cite{zhou2021transfill, xia2020joint, wang2019underexposed}. Similar to HDRNet, iHDRNet learns region-specific and intensity-specific color transformation parameters, which are applied to transform each input image adaptively. 
After color transformation, we expect that the domain discrepancy between inharmonious region and background could be magnified, so that the region localization network can identify the inharmonious region more easily. With this purpose, we leverage encoder to extract the domain-aware codes from inharmonious region and background before and after color transformation, in which the domain-aware codes are expected to contain the color and illumination statistics. Then, we design a Domain Discrepancy Magnification (DDM) loss to ensure that the distance of domain-aware codes between inharmonious region and background becomes larger after color transformation. Furthermore, we employ a Direction Invariance (DI) loss to regularize the domain-aware codes. 
For inharmonious region localization network, we can choose any existing network for region localization and place it under our framework. We refer to our framework as MadisNet (\textbf{Ma}gnifying \textbf{d}omain d\textbf{is}crepancy). 

We conduct experiments on the benchmark dataset iHarmony4~\cite{cong2020dovenet}, which shows that our proposed framework outperforms DIRL~\cite{liang2021inharmonious} and the state-of-the-art methods from other related fields. Our contributions can be summarized as follows:
\begin{itemize}
    \item We devise a simple yet effective inharmonious region localization framework which can accommodate any region localization method.
    \item We are the first to introduce adaptive color transformation to inharmonious region localization, in which improve HDRNet is used as the color mapping module.
    \item We propose a novel domain discrepancy magnification loss to magnify the domain discrepancy between inharmonious region and background.
    \item Extensive experiments demonstrate that our framework outperforms existing methods by a large margin (\emph{e.g.}, IoU is improved from 67.85\% to 74.44\%).
\end{itemize}

\section{Related Works}

\subsection{Image Harmonization}
Image harmonization, which aims to adjust the appearance of foreground to match background, is a long-standing research topic in computer vision. Prior works~\cite{cohen2006color, sunkavalli2010multi, jia2006drag, perez2003poisson, tao2010error} focused on transferring low-level appearance statistics from background to foreground. Recently, plenty of end-to-end solutions~\cite{tsai2017deep,cong2020dovenet,ling2021region,guo2021intrinsic,sofiiuk2021foreground} have been developed for image harmonization, including the first deep learning method~\cite{tsai2017deep}, domain translation based methods ~\cite{cong2020dovenet,cong2021bargainnet},  attention based module~\cite{cun2020improving,hao2020image}. Unfortunately, most of them require inharmonious region mask as input, otherwise the quality of harmonized image will be remarkably degraded. S2AM~\cite{cun2020improving} took blind image harmonization into account and predicted inharmonious region mask. However, mask prediction is not the focus of \cite{cun2020improving} and the quality of predicted masks is very low. 

\subsection{Inharmonious Region Localization}

Inharmonious region localization aims to spot the suspicious regions incompatible with background, from the perspective of color and illumination inconsistency. DIRL~\cite{liang2021inharmonious} was the first work on inharmonious region localization, which utilized bi-directional feature integration, mask-guided dual attention, and global-context guided decoder to dig out inharmonious regions.
Nevertheless, DIRL did not consider the uniqueness of this task and its performance awaits further improvement. 
In this work, we propose a novel framework to magnify the discrepancy between inharmonious region and background, which can help the downstream detector distinguish the inharmonious region from background. 

\subsection{Image Manipulation Localization}
Another related topic is image manipulation localization, which targets at distinguishing the tampered region from the pristine background. Copy-move, image splicing, removal, and enhancement are the four well-studied types in image manipulation localization, in which image splicing is the most related topic to our task.

Traditional image manipulation localization methods  heavily relied on the prior knowledge or strong assumptions on the inconsistency between tampered region and background, such as noise patterns~\cite{pun2016multi}, Color Filter Array interpolation patterns~\cite{ferrara2012image}, and JPEG-related compression artifacts~\cite{amerini2014splicing}. Recently, 
deep learning based methods~\cite{wu2019mantra, bappy2019hybrid, kniaz2019point, yang2020constrained}  attempted to tackle the image forgery problem by leveraging local patch comparison~\cite{bayar2016deep, rao2016deep, huh2018fighting, bappy2019hybrid}, forgery feature extraction~\cite{yang2020constrained, wu2019mantra,zhou2020generate}, adversarial learning~\cite{kniaz2019point}, and so on. Different from the above image manipulation localization methods,  color and illumination inconsistency is the main focus in inharmonious region localization task.

\subsection{Learnable Color Transformation}
In previous low-level computer vision tasks such as image enhancement, many color mapping techniques have been well explored, which meet our demand for color space manipulation. To name a few, HDRNet~\cite{gharbi2017deep} learned a guidance map and a bilateral grid to perform instance-aware linear color transformation.  Zeng \emph{et al.}~\cite{zeng2020learning} exploited 3D Look Up Table (LUT) for color transformation. DCENet~\cite{guo2020zero} iteratively estimated color curve parameters to correct color. In this work, we adopt the improved version of HDRNet~\cite{gharbi2017deep} as color mapping module to magnify the domain discrepancy between inharmonious region and background.

\section{Our Approach}
\begin{figure*}[!ht]
\centering
\includegraphics[width=1.0\linewidth]{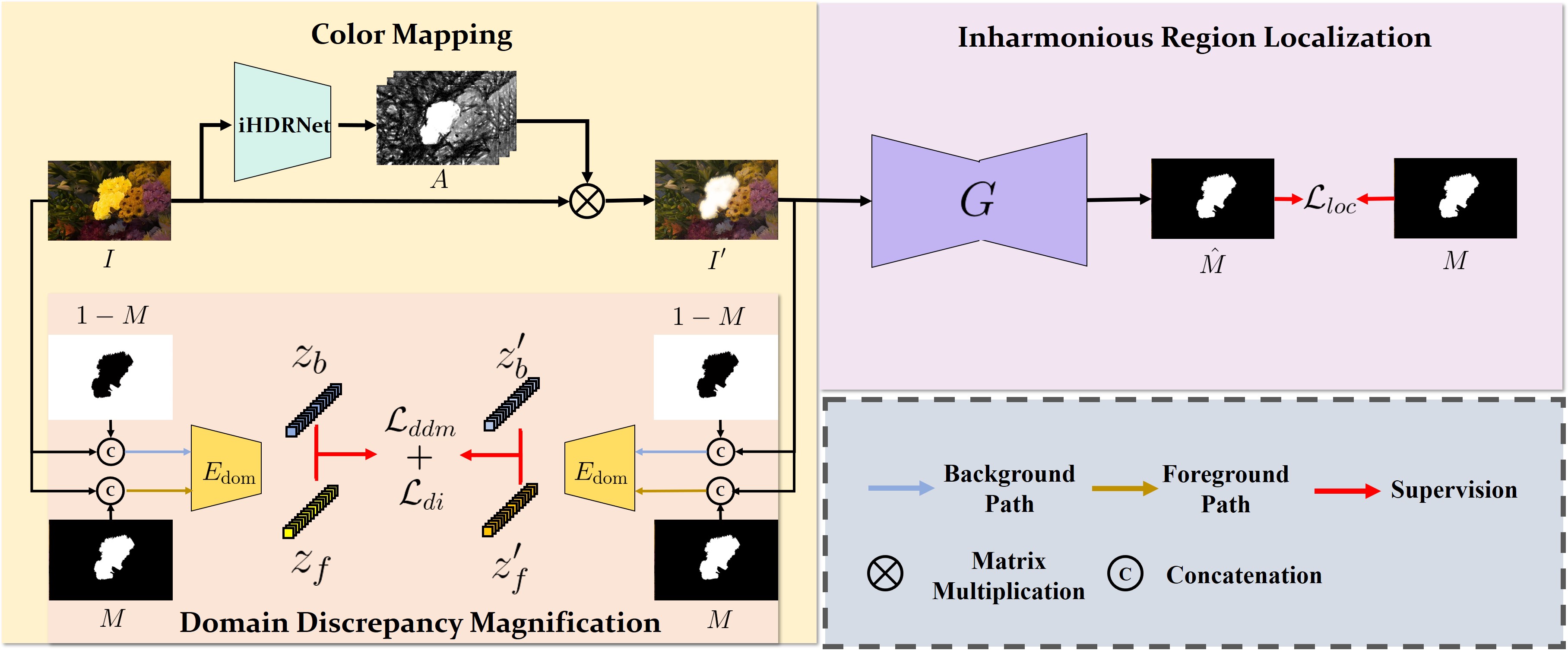} 
\caption{The illustration of our proposed framework which consists of color mapping stage and inharmonious region localization stage. Our color mapping module iHDRNet predicts the color transformation coefficients $\bm{A}$ for the input image $\bm{I}$, and the transformed image $\bm{I}'$ is fed into $G$ to produce the inharmonious region mask $\hat{\bm{M}}$. }
\label{fig:framework}
\end{figure*}

Given an input synthetic image $\bm{I}$, inharmonious region localization targets at predicting a mask $\hat{\bm{M}}$ that distinguishes the inharmonious region from the background region. Since the perception of inharmonious region is attributed to color and illumination inconsistency, we expect to find a color mapping $\mathcal{F}: \bm{I} \mapsto \bm{I}'$ so that the downstream localization network $G$ can capture the discrepancy between  inharmonious region and background more easily. As shown in Figure~\ref{fig:framework}, the whole framework consists of two stages: color mapping stage and inharmonious region localization stage. In the color mapping stage, we derive color transformation coefficients $\bm{A}$ from the color mapping module and perform color transformation to synthetic image $\bm{I}$ to produce the retouched image $\bm{I}'$. We assume that the retouched image $\bm{I}'$ will be exposed larger discrepancy between the inharmonious region and the background. To impose this constraint, we propose a domain discrepancy magnification loss and a direction invariance loss based on the extracted domain-aware codes of inharmonious regions and background regions in $\bm{I}$ and $\bm{I}'$. In the inharmonious region localization stage, the retouched image $\bm{I}'$ is delivered to the localization network $G$ to spot the inharmonious region, yielding the inharmonious mask $\hat{\bm{M}}$.  We will detail two stages in Section~\ref{sec:color_mapping} and Section~\ref{sec:localization_network} respectively. 

\subsection{Color Mapping Stage} \label{sec:color_mapping}
\subsubsection{Color Manipulation:}
In some localization tasks~\cite{panzade2016copy, roy2013face, cho2016canny, beniak2008automatic}, input images are first converted from RGB color space to other color spaces (\emph{e.g.}, HSV~\cite{panzade2016copy, roy2013face}, YCrCb~\cite{cho2016canny, beniak2008automatic}), in which the chroma and illumination distribution are more easily characterized. However, these color mappings are prefixed and cannot satisfy the requirement of inharmonious region localization task. Therefore, we seek to learn an instance-aware color mapping $\mathcal{F}: \bm{I} \mapsto \bm{I}'$, to promote the learning of downstream localization network. Considering the popularity of HDRNet~\cite{gharbi2017deep} and its remarkable success in color manipulation task~\cite{zhou2021transfill, xia2020joint, wang2019underexposed}, we build our color mapping module inheriting the spirits of HDRNet.
HDRNet~\cite{gharbi2017deep} implements local and global feature integration to keep texture details, producing a bilateral grid. 
To preserve edge information, they also learn an intensity map named guidance map and perform data-dependent lookups in the bilateral grid to generate region-specific and intensity-specific color transformation coefficients.
For more technique details, please refer to \cite{gharbi2017deep}.

We make two revisions for HDRNet. First, we first use central difference convolution layers~\cite{yu2020searching} to extract local features, in which a hyperparameter $\theta$ tradeoffs the contribution between vanilla convolution and central difference convolution.  As claimed in \cite{yu2020searching}, introducing central difference convolution into vanilla convolution can enhance the generalization ability and modeling capacity. Then, we apply a self-attention layer \cite{zhang2019self} to aggregate global information, which is adept at capturing long-range dependencies between distant pixels.
We use the processed features to produce the bilateral grid and the remaining steps are the same as HDRNet. We refer to the improved HDRNet as iHDRNet. The detailed comparison between HDRNet and iHDRNet can be found in Supplementary. 

Analogous to HDRNet, iHDRNet learns region-specific and intensity-specific color transformation coefficients $\bm{A} = [\bm{K}, \bm{b}] \in \mathbb{R}^{H\times W \times 3 \times 4}$ with $\bm{K}\in \mathbb{R}^{H\times W \times 3 \times 3}$ and $\bm{b}\in \mathbb{R}^{H\times W \times 3 \times 1}$, where $H$ and $W$ are the height and width of input image $\bm{I}$ respectively. 
    With color transformation coefficients $\bm{A}$, the inharmonious image $\bm{I}$ could be mapped to the retouched image $\bm{I}'$. Formally, for each pixel at location $p$, $\bm{I}'(p) = \bm{A}(p) \cdot [\bm{I}(p), \textbf{1}]^T = \bm{K}(p)\bm{I}(p) + \bm{b}(p)$, where $\bm{K}(p) \in \mathbb{R}^{3\times 3}, \bm{b}(p) \in \mathbb{R}^{3 \times 1}$ are the transform coefficients at location $p$. 

\subsubsection{Domain Discrepancy Magnification:}\label{sec:ddm}

We expect that the color and illumination discrepancy between the inharmonious region and the background is enlarged after color transformation. Following~\cite{cong2020dovenet,cong2021bargainnet}, we refer to each suite of color and illumination statistics as one domain. Then, we employ a domain encoder $E_\text{dom}$ to extract the domain-aware codes of inharmonious region and background separately from $\bm{I}$ and $\bm{I}'$. Note that we name the extracted code as domain-aware code instead of domain code, because the extracted code is expected to contain the color/illumination statistics but may also contain the content information (\emph{e.g.}, semantic layout). 
For the latent feature space, we select the commonly used intermediate features from the fixed pre-trained VGG-19~\cite{simonyan2014very} and pack them into the partial convolution layer~\cite{Liu2018} to derive region-aware features.   The domain encoder takes an image and a mask as input. Each partial convolutional layer performs convolution operation only within the masked area, where the mask is updated by rule and the information leakage from the unmasked area is avoided. At the end of $E_\text{dom}$, features are averaged along spatial dimensions and projected into a shape-independent domain-aware code. We denote the domain-aware codes of inharmonious region (\emph{resp.}, background) of $\bm{I}$ as $\bm{z}_f$ (\emph{resp.}, $\bm{z}_b$). Similarly, we denote the domain-aware code of inharmonious region (\emph{resp.}, background) of $\bm{I}'$ as $\bm{z}_f'$ (\emph{resp.}, $\bm{z}_b'$). Note that the domain encoder $E_\text{dom}$ is only used in the training phase, and only the projector is trainable while other components are frozen.

\subsubsection{Domain Discrepancy Magnification Loss:} To ensure that the color/illumination discrepancy between inharmonious region and background is enlarged, we enforce the distance between the domain-aware codes of inharmonious region and background of retouched image $\bm{I}'$ to be larger than that of original image $\bm{I}$. To this end, we propose a novel Domain Discrepancy Magnification (DDM) loss as follows,
\begin{eqnarray} \label{eqn:loss_ddm}
    \mathcal{L}_{{ddm}} = \max{(d(\bm{z}_f, \bm{z}_b) - d({\bm{z}'_f},{\bm{z}'_b}) + m,0)},
\end{eqnarray}
where $d(\cdot, \cdot)$ measures the Euclidean distance between two domain-aware codes, and the margin $m$ is set as $0.01$ via cross-validation. In this way, the distance between $\bm{z}'_f$ and $\bm{z}'_b$ is enforced to be larger than the distance between $\bm{z}_f$ and $\bm{z}_b$ by a margin $m$. One issue is that the domain-aware codes may also contain content information (\emph{e.g.}, semantic layout). However, the content difference between inharmonious region and background remains unchanged after color transformation, so we can deem $d(\bm{z}_f, \bm{z}_b) - d({\bm{z}'_f},{\bm{z}'_b})$ as the change in domain difference after color transformation. 

\subsubsection{Direction Invariance Loss: } In practice, we find that solely using (\ref{eqn:loss_ddm}) might lead to the corruption of domain-aware code space without necessary regularization. Inspired by StyleGAN-NADA~\cite{gal2021stylegannada}, we calculate the domain discrepancy vector $\Delta \bm{z} = \bm{z}_f - \bm{z}_b$ (\emph{resp.}, $\Delta \bm{z}' = \bm{z}_f' - \bm{z}_b'$) between inharmonious region and background in the input (\emph{resp.}, retouched) image. Then, we align the direction of domain discrepancy vector of input image with that of retouched image, using the following Direction Invariance (DI) loss:
\begin{eqnarray} \label{eqn:loss_di}
\begin{aligned}
    \mathcal{L}_{{di}} = 1 - \langle \Delta \bm{z},\Delta \bm{z}'  \rangle,
\end{aligned}
\end{eqnarray}
where $\langle \cdot, \cdot \rangle$ means the cosine similarity. Intuitively, we expect that the direction of domain discrepancy roughly stays the same after color transformation. There could be some other possible regularizers for domain-aware codes, but we observe that Direction Invariance (DI) loss in (\ref{eqn:loss_di}) empirically works well.

\subsection{Inharmonious Region Localization Stage} \label{sec:localization_network}
In the inharmonious region localization stage, the retouched image $\bm{I}'$ is delivered to the localization network $G$, which can dig out the inharmonious region from $\bm{I}'$ and produce the inharmonious mask $\hat{\bm{M}}$. 

The focus of this paper is a novel inharmonious region localization framework by magnifying the domain discrepancy. This framework can accommodate an arbitrary localization network $G$, such as inharmonious region localization method DIRL~\cite{liang2021inharmonious},  segmentation methods \cite{ronneberger2015u,chen2017rethinking}, and so on. In our experiments, we try using DIRL \cite{liang2021inharmonious} and UNet \cite{ronneberger2015u} as the localization network. 

After determining the region localization network, we wrap up its original loss terms (\emph{e.g.}, binary-cross entropy loss, intersection over union loss) as a localization loss $\mathcal{L}_{loc}$.
Together with our proposed domain discrepancy magnification (DDM) loss in (\ref{eqn:loss_ddm}) and direction invariance (DI) loss in (\ref{eqn:loss_di}), the total loss of our framework could be written as 
\begin{eqnarray}\label{eqn:total}
        \mathcal{L}_{total} = \lambda_{ddm} \mathcal{L}_{{ddm}} + \lambda_{di} \mathcal{L}_{di} + \mathcal{L}_{loc},
  \label{eqn:total}
\end{eqnarray}
where the trade-off parameter $\lambda_{ddm}$ and $\lambda_{di}$ depend on the downstream localization network.

\section{Experiments}

\subsection{Datasets and Implementation Details}
Following \cite{liang2021inharmonious}, we conduct experiments on the image harmonization dataset iHarmony4~\cite{cong2020dovenet}, which provides inharmonious images with their corresponding inharmonious region masks. iHarmony4 is composed of four sub-datasets: HCOCO, HFlickr, HAdobe5K,  HDay2Night. For HCOCO and HFlickr datasets, the inharmonious images are obtained by adjusting the color and lighting statistics of foreground. For HAdobe5K and HDay2Night datasets, the inharmonious images are obtained by overlaying the foreground with the counterpart of the same scene retouched with a different style or captured in a different condition. Therefore, the inharmonious images of the four sub-datasets will give people inharmonious perception mainly due to color and lighting inconsistency, which conforms to our definition of the inharmonious region.
Moreover, suggested by DIRL~\cite{liang2021inharmonious},  we simply discard the images with foreground occupying larger than 50\% area, which avoids the ambiguity that background can also be deemed as inharmonious region. 
Following \cite{liang2021inharmonious}, the training set and test set are tailored to 64255 images and 7237 images respectively. 

All experiments are conducted on a workstation with an Intel Xeon 12-core CPU(2.1 GHz), 128GB RAM, and a single Titan RTX GPU. We implement our method using Pytorch~\cite{paszke2019pytorch} with CUDA v10.2 on Ubuntu 18.04 and set the input image size as $256\times 256$. We choose Adam optimizer~\cite{kingma2014adam} with the initial learning rate 0.0001, batch size 8, and momentum parameters $\beta_1 = 0.5, \beta_2=0.999$. The hyper-parameter $\lambda_{ddm}$ and $\lambda_{di}$ in Eqn. (\ref{eqn:total}) are set as $0.01$ for DIRL\cite{liang2021inharmonious} and $0.001$ for UNet~\cite{ronneberger2015u} respectively. \emph{The detailed network architecture of domain encoder and iHDRNet can be found in Supplementary.}

For quantitative evaluation, we calculate Average Precision (AP), $F_1$ score, and Intersection over Union (IoU) based on the predicted mask $\hat{M}$ and the ground-truth mask $M$ following~\cite{liang2021inharmonious}.

\subsection{Baselines}
To the best of our knowledge, DIRL~\cite{liang2021inharmonious} is the only existing method designed for inharmonious region localization method. Therefore, we also consider other works from related fields. 1) blind image harmonization method S2AM~\cite{cun2020improving}; 2) image manipulation detection methods: MantraNet~\cite{wu2019mantra}, MFCN~\cite{salloum2018image}, MAGritte~\cite{kniaz2019point}, H-LSTM~\cite{bappy2019hybrid}, SPAN~\cite{hu2020span}; 3) salient object detection methods: F3Net~\cite{wei2020f3net}, GATENet~\cite{zhao2020suppress}, MINet~\cite{pang2020multi}; 4) semantic segmentation methods: UNet~\cite{ronneberger2015u}, DeepLabv3~\cite{chen2017rethinking}, HRNet-OCR~\cite{sun2019deep}. 

\begin{table}[t] 
\centering
  \begin{tabular} {l|c|c|c }
    \toprule[1pt]
\multirow{2}{1.0in}{\textbf{Methods}} &  \multicolumn{3}{c}{\textbf{Evaluation Metrics}} \\ \cline{2-4}
                 & \multicolumn{1}{c|}{AP(\%) $\uparrow$}  &  \multicolumn{1}{c|}{$F_1$ $\uparrow$} & \multicolumn{1}{c}{IoU(\%) $\uparrow$} 
    \\    \hline 
   \textbf{UNet} &  74.90 & 0.6717 & 64.74 \\ 
   \textbf{DeepLabv3} & 75.69 & 0.6902 & 66.01 \\
   \textbf{HRNet-OCR} & 75.33 & 0.6765 & 65.49 \\ \hline
   \textbf{MFCN} & 45.63 & 0.3794 & 28.54 \\
    \textbf{MantraNet}  & 64.22 & 0.5691 & 50.31 \\
     \textbf{MAGritte}  & 71.16 & 0.6907 & 60.14 \\
      \textbf{H-LSTM}  & 60.21 & 0.5239 & 47.07 \\ 
       \textbf{SPAN}  & 65.94 & 0.5850 & 54.27 \\ \hline
       \textbf{F3Net}  & 61.46 & 0.5506 & 47.48 \\
       \textbf{GATENet}  & 62.43 & 0.5296 & 46.33 \\
       \textbf{MINet}  & 77.51 & 0.6822 & 63.04 \\ \hline
       \textbf{S2AM}    & 43.77  & 0.3029 & 22.36 \\ \hline
       \textbf{DIRL} & 80.02 & 0.7317 & 67.85 \\ \hline
        \textbf{MadisNet(UNet)} & 81.15 & 0.7372 & 67.28 \\  
       \textbf{MadisNet(DIRL)}  & \textbf{85.86} & \textbf{0.8022} & \textbf{74.44} \\  
    \bottomrule[1pt]
  \end{tabular}
\caption{Quantitative comparison with baseline methods on iHarmony4 dataset. The best results are denoted in boldface.}
\label{tab:quantitative}
\end{table}

\begin{figure*}[!ht]
\centering
\includegraphics[width=0.95\linewidth]{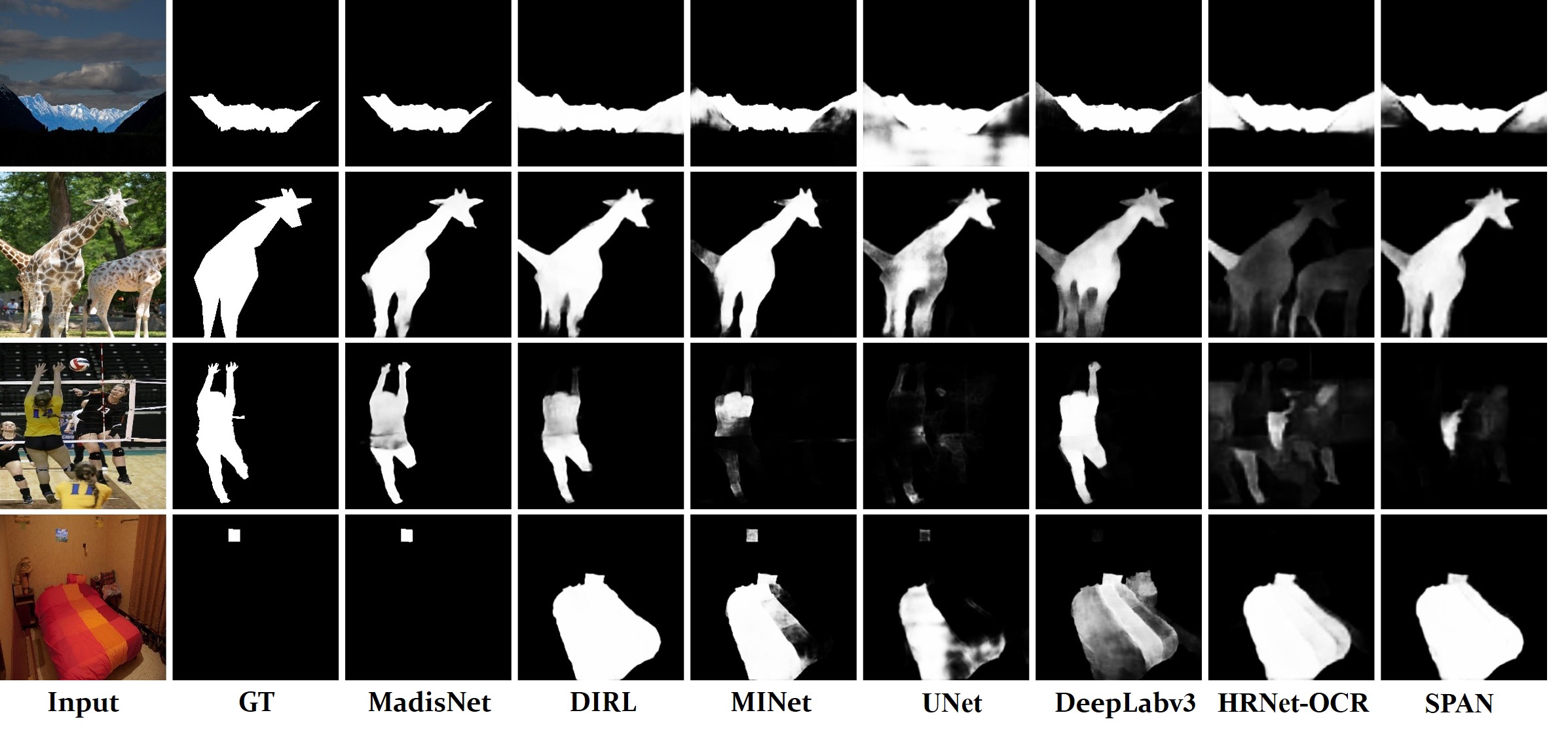} 
\caption{Qualitative comparison with baseline methods. GT is the ground-truth inharmonious region mask. }
\label{fig:qualitative}
\end{figure*}

\subsection{Experimental Results}
\subsubsection{Quantitative Comparison}
The quantitative results are summarized in Table~\ref{tab:quantitative}. All of the baseline results are directly copied from \cite{liang2021inharmonious} except  SPAN, GATENet, F3Net, and MINet. For fair comparison, we trained the baselines from scratch. 
One observation is that image manipulation localization methods~\cite{wu2019mantra, kniaz2019point, bappy2019hybrid, hu2020span} are weak in localizing the inharmonious region. One possible explanation is that they focus on the noise pattern and forgery feature extraction while paying less attention to the low-level statistics of color and illumination. We also notice that salient object detection methods~\cite{wei2020f3net, zhao2020suppress} also achieve  worse performance than the semantic segmentation methods~\cite{ronneberger2015u, chen2017rethinking, sun2019deep} while MINet~\cite{pang2020multi} beats all of the semantic segmentation methods in AP metric.
In S2AM~\cite{cun2020improving}, they predict an inharmonious region mask as side product to indicate the region to be harmonized. Unfortunately, the quality of inharmonious mask is far from satisfactory since image harmonization is their main focus. 
Another interesting observation is that typical segmentation methods achieve the most competitive performance among the methods that are not specifically designed for inharmonious region localization. It might be attributed to that semantic segmentation methods are originally designed in a general framework and generalizable to inharmonious region localization task.

Since our framework can accommodate any region localization network, we explore using UNet and DIRL under our framework, which are referred to as MadisNet(UNet) and MadisNet(DIRL) respectively.  
It can be seen that MadisNet(DIRL) (\emph{resp.}, MadisNet(UNet)) outperforms DIRL (\emph{resp.}, UNet).
MadisNet(DIRL) beats the existing inharmonious region localization method and all of the state-of-the-art methods from related fields by a large margin, which verifies the effectiveness of our framework. \emph{In the remainder of experiment section, we use DIRL as our default region localization network (i.e., ``MadisNet" is short for ``MadisNet(DIRL)"), unless otherwise specified. }

\subsubsection{Qualitative Comparison}
We show the visualization results as well as baselines in Figure~\ref{fig:qualitative}, which shows that our method can localize the inharmonious region correctly and preserve the boundaries accurately.  In comparison, the baseline methods may locate the wrong object (row 4) or only detect an incomplete region (row 3). More visualization results can be found in Supplementary.

\subsection{Ablation Studies}
\subsubsection{Loss Terms}
First, we analyze the necessity of each loss term in Table~\ref{tab:losses}. One can learn that our proposed $\mathcal{L}_{ddm}$ and $\mathcal{L}_{di}$ are complementary to each other. Without our proposed $\mathcal{L}_{ddm}$ and $\mathcal{L}_{di}$, the performance is significantly degraded, which proves that $\mathcal{L}_{ddm}$ and  $\mathcal{L}_{di}$ play important roles in inharmonious region localization.

\begin{table}[t] 
\centering
 \begin{tabular} {c|c | c|c|c }
    \toprule[1pt]
\multicolumn{2}{c|}{\textbf{Components}} & \multicolumn{3}{c}{\textbf{Evaluation Metrics}}\\ \cline{1-5}
                \multicolumn{1}{c|}{Encoder }  &  
                \multicolumn{1}{c|}{Self Attention} &
                \multicolumn{1}{c|}{AP $\uparrow$} &
                \multicolumn{1}{c|}{$F_1$ $\uparrow$} &
                \multicolumn{1}{c}{IoU $\uparrow$}  \\
     \hline 
    VC  & &  81.05 & 0.7508 & 69.43  \\ 
    VC & \checkmark  &  83.54 & 0.7749 & 72.08 \\ 
    CDC &  &  82.80 & 0.7697 & 71.64 \\ 
    CDC & \checkmark  &  \textbf{85.86} & \textbf{0.8022} & \textbf{74.44} \\ 
    \bottomrule[1pt]
 \end{tabular}
\caption{Ablation study on the components of improved HDRNet. ``VC" denotes the vanilla convolution layer and ``CDC" means the central difference convolution layer.}
\label{tab:iHDRNet}
\end{table}

\subsubsection{iHDRNet}
Then, we conduct ablation study to validate the effectiveness of CDC layer and self-attention layer in our iHDRNet. The results are summarized in Table~\ref{tab:iHDRNet}. By comparing row 1 (\emph{resp.}, 3) and row 2 (\emph{resp.}, 4), we can see that it is useful to employ self-attention layer to capture the long-range dependencies with promising improvement. The comparison between row 2 and row 4 demonstrates that CDC layer performs more favorably than vanilla convolution layer, since CDC layer can capture both intensity-level information and gradient-level information.  

\begin{table}[t] 
\centering
  \begin{tabular} {l |c|c|c }
    \toprule[1pt]
\multirow{2}{0.8in}{\textbf{Loss Terms}} &  \multicolumn{3}{c}{\textbf{Evaluation Metrics}} \\ \cline{2-4}
                 & \multicolumn{1}{c|}{AP(\%) $\uparrow$}  &  \multicolumn{1}{c|}{$F_1$ $\uparrow$} & \multicolumn{1}{c}{IoU(\%) $\uparrow$} 
    \\    \hline
    $\mathcal{L}_{loc}$ & 80.95 & 0.7401 & 68.81 \\ \hline
    $\mathcal{L}_{loc} + \mathcal{L}_{ddm}$ &  81.86 & 0.7533 & 69.84 \\ 
    $\mathcal{L}_{loc} + \mathcal{L}_{di}$ &  83.18 & 0.7701 & 71.67 \\ 
  $\mathcal{L}_{loc} + \mathcal{L}_{ddm} + \mathcal{L}_{di}$ &  \textbf{85.86} & \textbf{0.8022} & \textbf{74.44}  \\ 
    \bottomrule[1pt]
  \end{tabular}
\caption{The comparison among different loss terms.}
\label{tab:losses}
\end{table}

\subsection{Study on Color Manipulation Approaches}

To find the best color manipulation approach for inharmonious region localization, we compare our color mapping module iHDRNet with non-learnable color transformation and learnable color transformation. 

For non-learnable color transformation, we transform the input RGB image to other color spaces (HSV, YCrCb). Besides, one might concern that whether the learnable color mapping is equivalent to applying random color jittering to the input image, thereby we also take the color jittering augmentation into account. Because the above color transformation approaches do not involve learnable model parameters, we simply apply them to the input images and feed transformed images into the region localization network, during which DDM loss and DI loss are not used.   

For learnable color transformation, we compare with  LUTs \cite{zeng2020learning}, DCENet~\cite{guo2020zero}, and HDRNet~\cite{gharbi2017deep}. 
We directly replace iHDRNet with these color transformation approaches and the other components of our proposed framework remain the same, in which DDM loss and DI loss are used.

The results are summarized in Table~\ref{tab:ddm_variants}. 
We also include the RGB baseline, which means that no color mapping is applied, and the result is identical with DIRL in Table~\ref{tab:quantitative}.
One can observe that the non-learnable color mapping methods achieve comparable or even worse results compared with RGB baseline. We infer that they are unable to reveal the relationship between inharmonious region and background through simple traditional color transformation. 
In learnable color mapping methods, LUT achieves even worse scores than RGB baseline. This might be that LUT only learns a global transformation for the whole image without considering local variation. 
HDRNet and DCENet slightly improve the performance. One possible explanation is that both HDRNet and DCENet are region-specific color manipulation methods, so they could learn color transformation for different regions adaptively to make downstream localization module easily discover the inharmonious region. 
Our iHDRNet achieves the best results, because the central difference convolution~\cite{yu2020searching} can help identify the color inconsistency in synthetic images and the self-attention layer can capture long-range dependencies between distant pixels.

\begin{table}[t] 
\centering
  \begin{tabular} {l |c|c|c }
    \toprule[1pt]
\multirow{2}{1.1in}{\textbf{Color Mapping}} &  \multicolumn{3}{c}{\textbf{Evaluation Metrics}} \\ \cline{2-4}
                 & \multicolumn{1}{c|}{AP(\%) $\uparrow$}  &  \multicolumn{1}{c|}{$F_1$ $\uparrow$} & \multicolumn{1}{c}{IoU(\%) $\uparrow$} 
    \\    \hline
    \textbf{RGB}(Baseline) & 80.02 & 0.7317 & 67.85 \\ \hline
    \textbf{HSV} &  79.86 & 0.7282 & 67.40 \\ 
    \textbf{YCrCb} &  81.07 & 0.7484 & 69.35 \\ 
   \textbf{ColorJitter} &  77.50 & 0.7068 & 65.40 \\ \hline
   \textbf{LUTs} &  78.39 & 0.7181 & 66.16 \\ 
     \textbf{DCENet} &  81.90 & 0.7623 & 70.92 \\ 
    \textbf{HDRNet} &  81.05 & 0.7508 & 69.43 \\ \hline
    \textbf{iHDRNet} &  \textbf{85.86} & \textbf{0.8022} & \textbf{74.44} \\ 
    \bottomrule[1pt]
  \end{tabular}
\caption{The comparison among different color mapping methods. RGB(baseline) means that no color mapping is applied.}
\label{tab:ddm_variants}
\end{table}

\subsection{Analyses of Domain Discrepancy}

\begin{table}[t] 
\centering
  \begin{tabular} {|c|c|c|} \hline
  & $d_{f,b} + m< d'_{f,b}$ & $d_{f,b} < d'_{f,b}$   \\ \hline 
 Training set & 76.22\% & 99.74\%  \\ \hline
 Test set & 77.38\% & 99.68\% \\ \hline
  \end{tabular}
\caption{The percentage of images whose domain discrepancy is enlarged after color mapping. $d_{f,b}$ is short for $d(\bm{z}_f,\bm{z}_b)$ and $d'_{f,b}$ is short for $d(\bm{z}'_f,\bm{z}'_b)$. Here $m = 0.01$ as described in section ~\ref{sec:ddm}.}
\label{tab:domain_discrepancy}
\end{table}

\begin{figure}[t]
\centering
\includegraphics[width=0.8\linewidth]{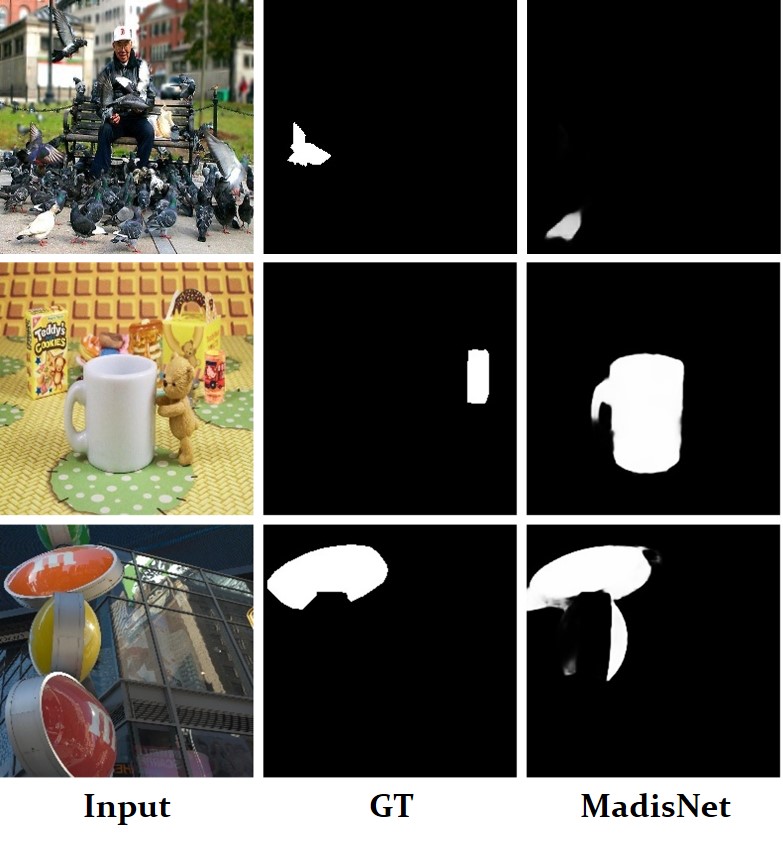} 
\caption{Failure cases of our method. ``GT" is the ground-truth inharmonious region mask.}
\label{fig:failure}
\end{figure}

We report the percentage of images whose domain discrepancy is magnified after color transformation in Table \ref{tab:domain_discrepancy}. For both training set and test set, we report two results: the percentage of $d(\bm{z}_f,\bm{z}_b)+m<d(\bm{z}'_f,\bm{z}'_b)$ and the percentage of $d(\bm{z}_f,\bm{z}_b)<d(\bm{z}'_f,\bm{z}'_b)$, in which the latter one is a special case of the former one by setting $m=0$. From Table \ref{tab:domain_discrepancy}, we can see that 
the color mapping module learnt on the training set can generalize to the test set very well. In test set, the domain discrepancy of $77.38\%$ images is enlarged by at least a margin $m$ after color transformation. When we relax the requirement, \emph{i.e.}, $m=0$, the percentage is as high as $99.68\%$ on the test set.

\subsection{Discussion on Limitation}

Figure~\ref{fig:failure} shows three failure cases of our model. In row 1, our model treats the white pigeon at the bottom left of image as the inharmonious region. We conjecture that the inharmonious region has similar dark tone with surrounding pigeons so that our model is misled by the white pigeon. In row 2, the white cup is recognized as the inharmonious region, probably because the ground-truth inharmonious region and background share warm color tone. In the last row, our model views the yellow light sign as  inharmonious region too, because the inharmonious region is brighter than the background. In summary, our model may be weak when the target inharmonious region is surrounded by objects with similar color or intensity.  

\subsection{Results on Four Sub-datasets and Multiple Inharmonious Regions}
Because iHarmony4~\cite{cong2020dovenet} contains four sub-datasets, we show the results on four sub-datasets in Supplementary.
Furthermore, this paper mainly focuses on one inharmonious region, but there could be multiple disjoint inharmonious regions in a synthetic image. Therefore, we also demonstrate the ability of our method to identify multiple disjoint inharmonious regions in Supplementary.

\section{Conclusion}
In this paper, we have proposed a novel framework to resolve the inharmonious region localization problem with color mapping module and our designed domain discrepancy magnification loss. With the process of color mapping module, the inharmonious region could be more easily discovered from the synthetic images. Extensive experiments on iHarmony4 dataset have demonstrated the effectiveness of our approach.

\section*{Acknowledgements}
This work is partially sponsored by National Natural Science Foundation of China (Grant No. 61902247),  Shanghai Municipal Science and Technology Major Project (2021SHZDZX0102), Shanghai Municipal Science and Technology Key Project (Grant No. 20511100300). 
\bibliographystyle{aaai22}
\bibliography{1.main.bbl}

\end{document}


\maketitle
In this document, we provide additional materials to support our main submission. In Section \ref{sec:implementation}, we introduce the details of our network architecture including the domain encoder and the improved HDRNet (iHDRNet). In Section~\ref{sec:hyperparameter}, we analyze the impact of hyperparameters $m$, $\lambda_{ddm}$, and $\lambda_{di}$. In Section~\ref{sec:visual}, we show more visualization results of our method and baseline methods. In Section~\ref{sec:color_local}, we investigate the impact of color transformation on localization results. Finally, we report the results on all four sub-datasets in Section~\ref{sec:four_datasets} and the results for multiple inharmonious regions in Section~\ref{sec:multiple_regions}. 

\begin{figure*}[ht]
\centering
\includegraphics[width=0.95\linewidth]{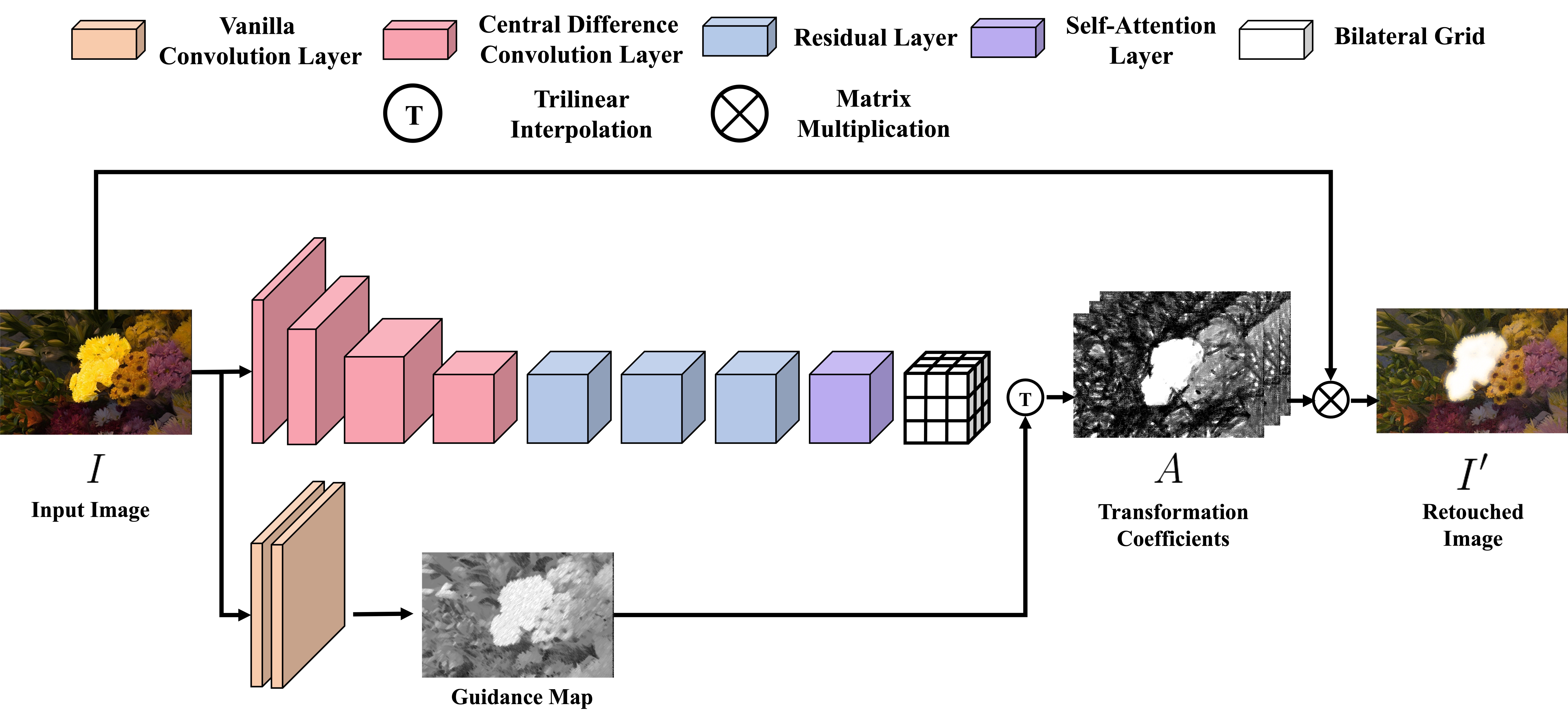} 
\caption{The architecture of improved HDRNet (iHDRNet). }
\label{fig:iHDRNet}
\end{figure*}

\section{Network Architecture}\label{sec:implementation}
In this section, we elaborate the details of domain-aware code extractor $E_\text{dom}$ and our improved HDRNet (iHDRNet). 
\subsection{Domain Encoder}
\begin{figure}[t]
\centering
\includegraphics[width=0.95\linewidth]{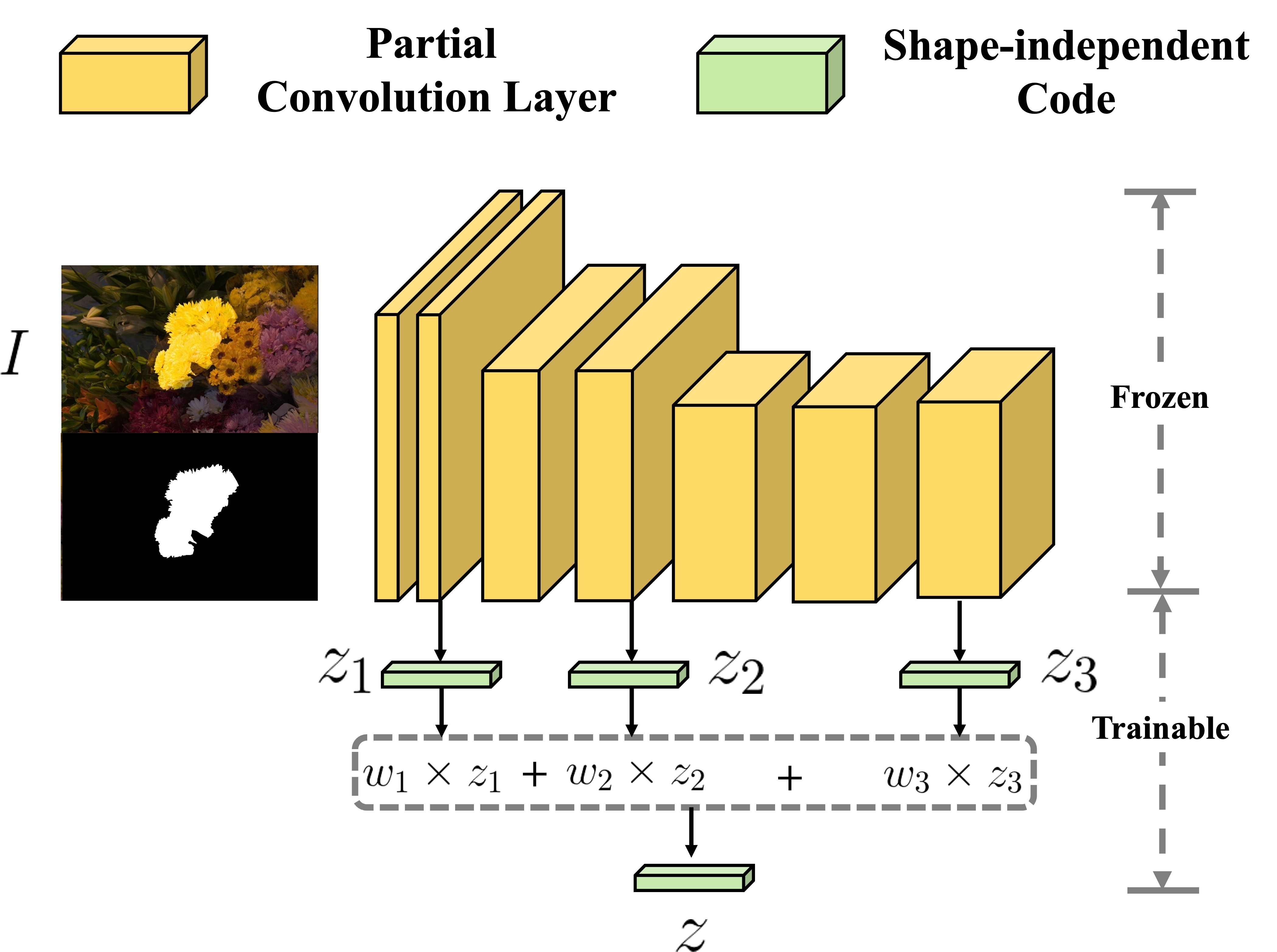} 
\caption{The architecture of domain encoder $E_\text{dom}$. }
\label{fig:E_dom}
\end{figure}

We employ a domain encoder $E_\text{dom}$ to extract the domain-aware codes of inharmonious region and background separately. The domain encoder takes an image and a mask as input. The network structure of domain encoder is shown in Figure~\ref{fig:E_dom}. 
We use pre-trained VGG-19~\cite{simonyan2014very} to extract multi-scale features, in which the vanilla convolution layer is replaced by partial convolution layer \cite{Liu2018}. Each partial convolutional layer performs convolution operation only within the masked area, where the mask is updated by rule and the information leakage from the unmasked area is prevented.
For the technical details of partial convolution, please refer to \cite{Liu2018}. 
We take the \texttt{conv1\_2}, \texttt{conv2\_2}, \texttt{conv3\_3} features, which separately pass through a global average pooling layer and a fully-connected layer to yield three shape-independent vectors $\{\mathbf{z}_i|^3_{i=1}\}, \mathbf{z}_i \in \mathbb{R}^{d_{dom}}$ with $d_{dom}$ being the latent code dimension. Here, we set $d_{dom} = 16$ following the configuration of BargainNet~\cite{cong2021bargainnet}. Finally,  we use learnable weights $\{w_i|^3_{i=1}\}$  combined with $\{\mathbf{z}_i|^3_{i=1}\}$ to produce the final domain-aware code $\mathbf{z}$, which can be formulated as $\mathbf{z} = \sum_{i=1}^3 w_i \cdot \mathbf{z}_i$.

\subsection{Improved HDRNet}

As shown in Figure~\ref{fig:iHDRNet}, we build up our improved HDRNet inheriting the spirits of HDRNet~\cite{gharbi2017deep}.
At first,  we replace the vanilla convolution used in encoder with central difference convolution layer~\cite{yu2020searching}. As indicated in CDCNet~\cite{yu2020searching}, Central Difference Convolution (CDC) layer merges central difference term into vanilla convolution layer, which can capture both intensity-level information and gradient-level information.  CDC layer combines the advantages of central difference term and vanilla convolution, which can enhance the generalization ability and modeling capacity.

In HDRNet~\cite{gharbi2017deep}, global features are acquired by global average pooling on local features. We argue that the extracted global features may not be capable of revealing the relation between regions, in which inharmonious regions are determined by comparing with other regions in an image. Therefore, we replace the global feature branch with self-attention layer~\cite{zhang2019self} to model the long-range dependence between distant pixels.
In self-attention layer, input feature map $\textbf{X} \in \mathbb{R}^{H\times W \times C}$ are projected into three terms named query $\textbf{Q} \in \mathbb{R}^{H\times W \times C'}$, key $\textbf{K} \in \mathbb{R}^{H\times W \times C'}$, and value $\textbf{V} \in \mathbb{R}^{H\times W \times C}$ respectively, where $H, W$ refer to spatial size and $C$ (\emph{resp.}, $C'$) means the original channel dimension (\emph{resp.}, reduced channel dimension). Then, the output feature $\textbf{Y}$ augmented by self-attention is computed by:
\begin{eqnarray}
\begin{aligned}
 \textbf{X}' = \text{Softmax}(\frac{\textbf{Q}\textbf{K}^{T}}{\sqrt{C'}})\cdot \textbf{V},\quad \textbf{Y} = \textbf{X} + \gamma \cdot \textbf{X}',
 \end{aligned}
\end{eqnarray}
where $\gamma$ is a learnable parameter that reweights the contribution of self-attention feature $\textbf{X}'$. Please refer to~\citet{zhang2019self} for more technical details.

\begin{figure*}[t]
\centering
\includegraphics[width=0.9\linewidth]{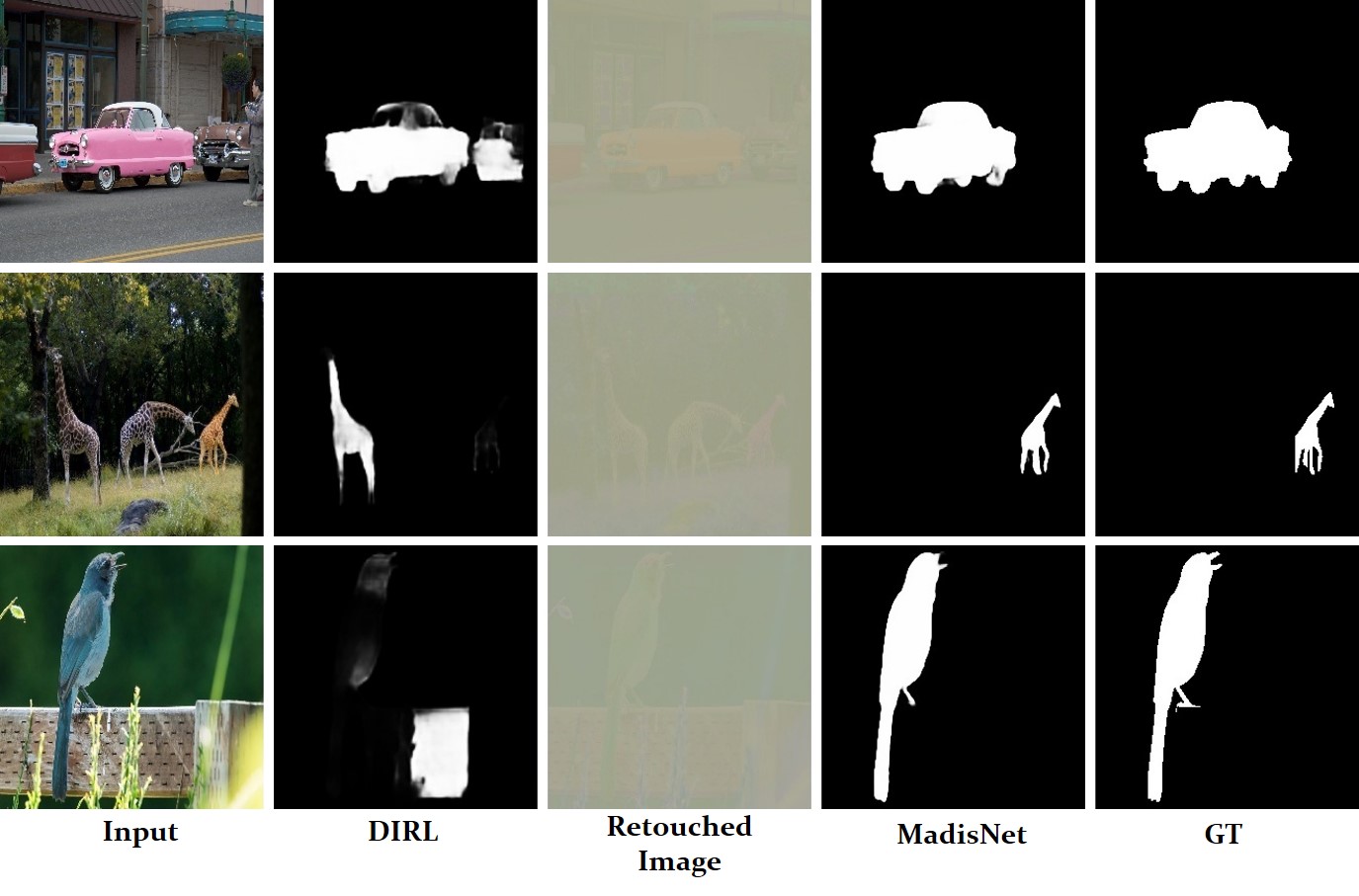} 
\caption{The impact of our color mapping module on localization results. ``Retouched Image" is the output of color mapping module. ``GT" is the ground-truth inharmonious region mask. }\label{sec:transform_cmp}
\label{fig:transform_cmp}
\end{figure*}

\subsection{Analysis of Model Size}
We report the model size of each module in our framework, including the domain encoder $E_\text{dom}$, our iHDRNet, and the region localization network DIRL~\cite{liang2021inharmonious}. The comparison among three modules shows that the extra modules (domain encoder and iHDRNet) added in our framework are relatively efficient. Moreover, the domain encoder is only used in the training stage, while not used in the testing stage. 

\begin{table}[t] 
\centering
  \begin{tabular} {c|c|c|c} 
  \toprule[1pt]
  Module & iHDRNet & $E_\text{dom}$ & DIRL \\ \hline
 Size(M) & 1.47  & 1.78  & 53.47 \\ 
  \bottomrule[1pt]
  \end{tabular}
\caption{The model size of each module in our method.}
\label{tab:model_size}
\end{table}

\section{Hyper-parameter Analyses}\label{sec:hyperparameter}
\begin{figure*}[t]
\centering
\includegraphics[width=1.0\linewidth]{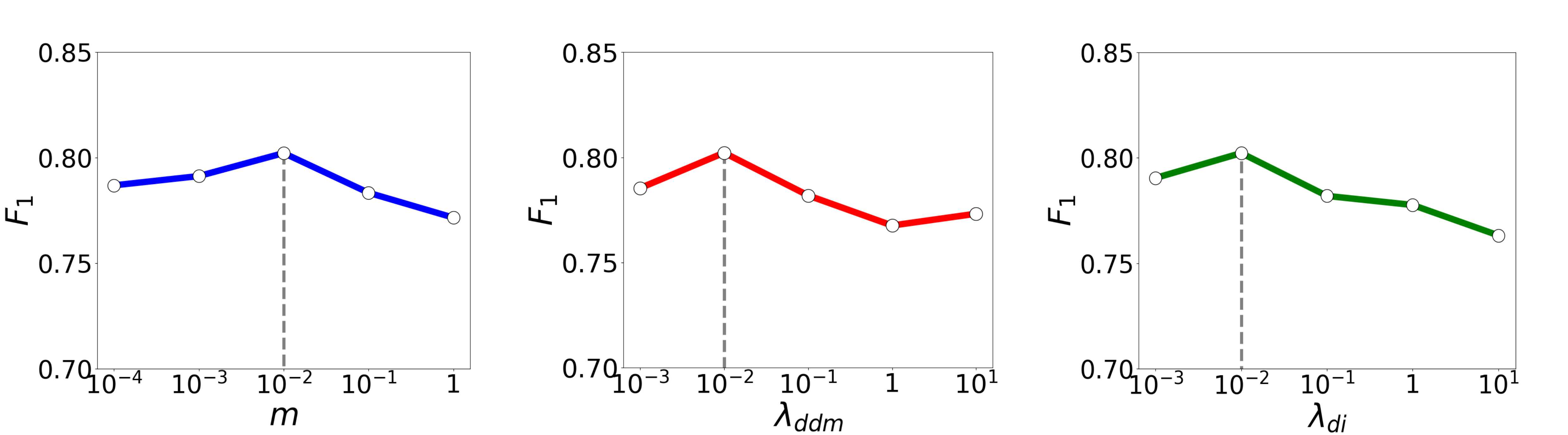} 
\caption{$F_1$ of our method when varying the hyper-parameters $m$, $\lambda_{ddm}$, and $\lambda_{di}$.}
\label{fig:hyperparamter}
\end{figure*}

\begin{table}[t] 
\centering
\setlength{\tabcolsep}{1.25mm}{
\scalebox{0.95}{
 \begin{tabular} {|c| c|c|c| c|c|c|} \hline
\textbf{Method} & \multicolumn{3}{c|}{\textbf{DIRL}} & \multicolumn{3}{c|}{\textbf{MadisNet(DIRL)}} \\ \hline
                \multicolumn{1}{|c|}{Metrics}  &  
                \multicolumn{1}{c|}{AP $\uparrow$ } &
                \multicolumn{1}{c|}{$F_1$ $\uparrow$} &
                \multicolumn{1}{c|}{IoU $\uparrow$} &
                \multicolumn{1}{c|}{AP $\uparrow$} &
                \multicolumn{1}{c|}{$F_1$ $\uparrow$} &
                \multicolumn{1}{c|}{IoU $\uparrow$}  \\ \hline 
  \text{HCOCO} & 74.25 & 0.6701 &  60.85 & 83.78 & 0.7741 &  70.50 \\ \hline
    \text{HAdobe5k} & 92.16 & 0.8801 & 84.02 & 92.45 & 0.8850 & 84.75 \\ \hline
    \text{HFlickr} & 84.21 & 0.7786 & 73.21 & 85.65 & 0.8032 & 75.49 \\ \hline
    \text{HDay2night} & 38.74 & 0.2396 & 20.11 & 57.40 & 0.4672 & 40.47 \\ \hline
 \end{tabular}
 }
 }
\caption{Quantitative comparison with baseline method DIRL on four sub-datasets of iHarmony4.}
\label{tab:sub_datasets}
\end{table}

\begin{figure}[t]
\centering
\includegraphics[width=0.95\linewidth]{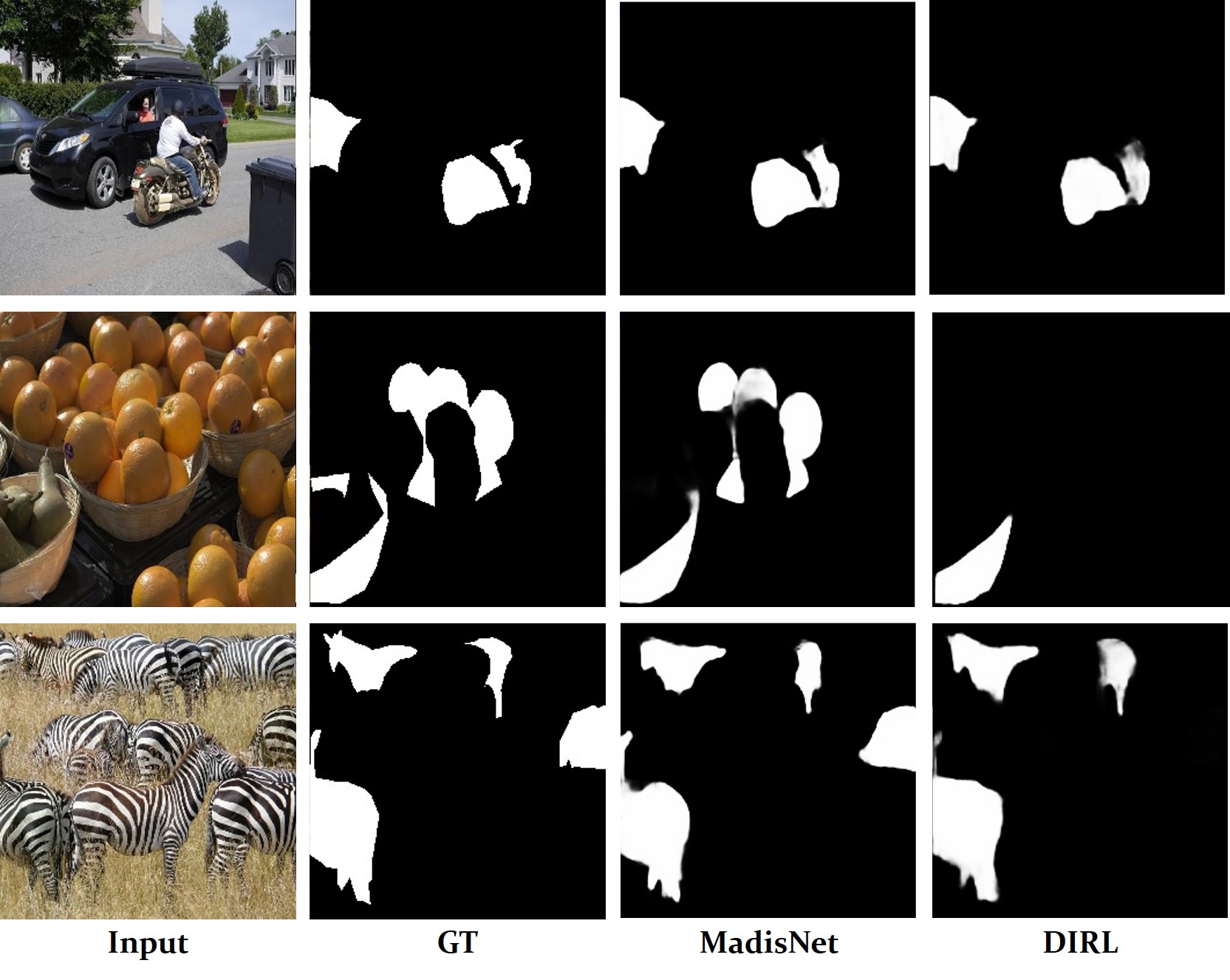} 
\caption{The visualization comparison on the synthetic images with multiple inharmonious regions.  ``GT" is the ground-truth inharmonious region mask.}
\label{fig:multi_objects}
\end{figure}

In this section, we investigate the impact of hyper-parameters $m$, $\lambda_{ddm}$, and $\lambda_{di}$. We vary $m$ (\emph{resp.}, $\lambda_{ddm}$, $\lambda_{di}$) in the range of [$10^{-4}$, $1$] (\emph{resp.}, [$10^{-3}$, $10^1$], [$10^{-3}$, $10^1$]) and plot $F_1$ of our method in  Figure \ref{fig:hyperparamter}, in which we tune the hyperparameter reported in the horizontal axis and keep others fixed. We use beam search for the hyperparameter $\{\lambda_{ddm},\lambda_{di}\}$ and then fixed the best $\{\lambda_{ddm},\lambda_{di}\}$ for hyperparameter $m$ configuration. It can be seen that our method is relatively robust with the hyper-parameters $m$,  $\lambda_{ddm}$, and $\lambda_{di}$ when setting them in a reasonable range.

\section{More Visualization Results}\label{sec:visual}
\begin{figure*}[t]
\centering
\includegraphics[width=1.0\linewidth]{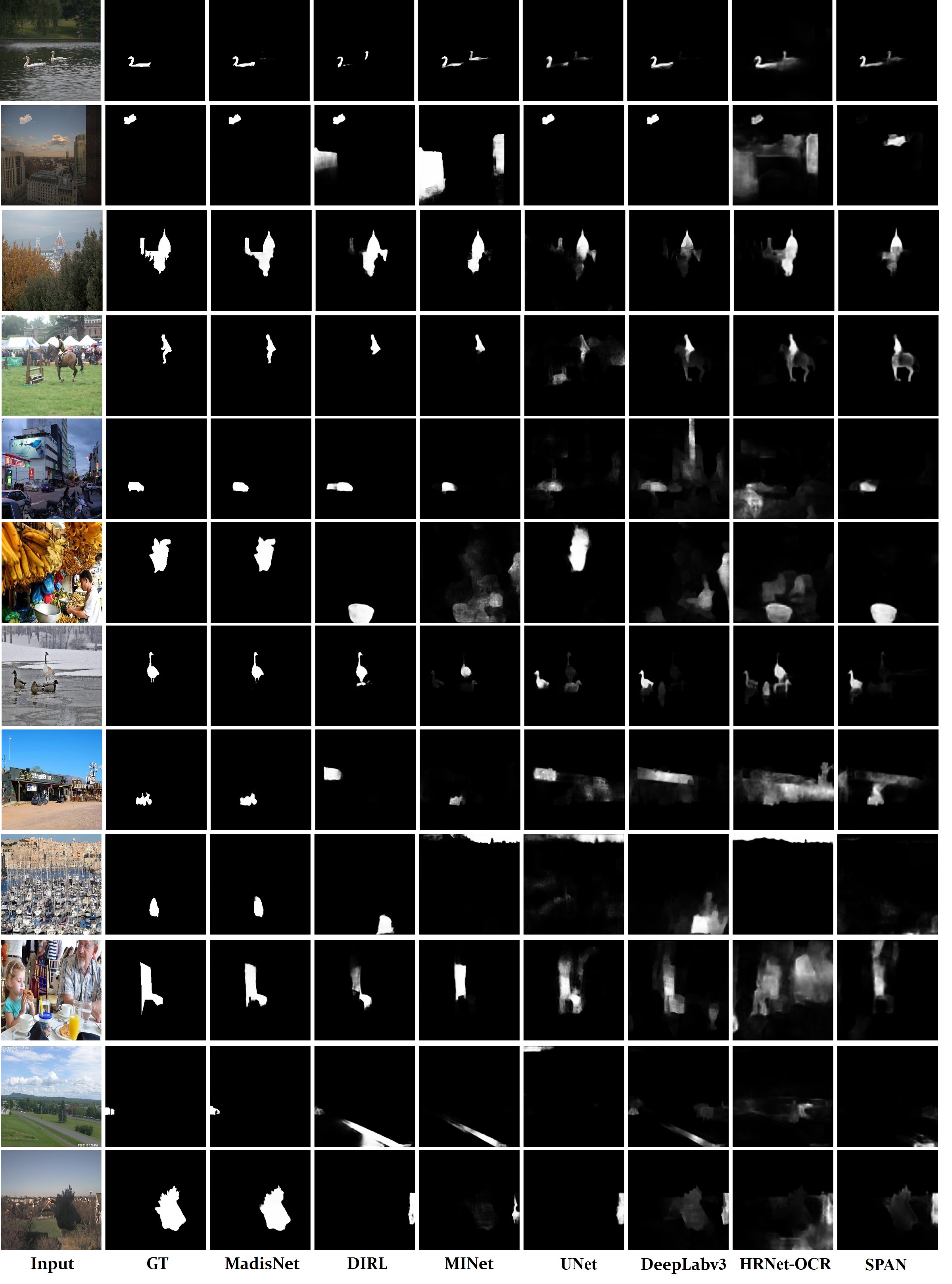} 
\caption{Qualitative comparison with baseline methods. GT is the ground-truth inharmonious region mask. }\label{sec:visulization}
\label{fig:visualization}
\end{figure*}

We show more examples of the comparison between our method and baseline methods DIRL~\cite{liang2021inharmonious}, MINet~\cite{pang2020multi}, UNet~\cite{ronneberger2015u}, Deeplabv3~\cite{chen2017rethinking}, HRNet-OCR~\cite{sun2019deep}, and SPAN~\cite{hu2020span} in Figure \ref{fig:visualization}. In row 1, we can see that our method successfully shows the right inharmonious swan, while most of the baselines are struggling to identify the inharmonious swan and even assign inharmonious label to another harmonious swan. In row 3, our method is capable of depicting the accurate boundary of inharmonious region and other baselines suffer from the shadow artifacts. In row 9, the input synthetic image is an extremely challenging scene, in which the inharmonious boat is surrounded by crowded boats. Our method is powerful enough to distinguish the inharmonious boat from the background, whereas other methods are unable to detect such region. These visualization results demonstrate the superiority of our method again.

\section{The Impact of Color Transformation on Localization Results}\label{sec:color_local}
In Figure~\ref{fig:transform_cmp}, we demonstrate the impact of our proposed color mapping module on the localization results. Here, ``DIRL" refers to results of the DIRL~\cite{liang2021inharmonious} model trained with RGB input directly. ``Retouched Image" is the output of color mapping module. ``MadisNet" means that the results are derived from the DIRL model trained with the output of color mapping module. From the localization results, we can see that in row 1, MadisNet removes the false alarm from which original DIRL suffers. In row 2 and 3, MadisNet localizes the inharmonious region correctly while original DIRL is distracted by other misleading regions. 

The improvement of MadisNet is attributed to the color mapping module which enlarges the domain discrepancy and makes the downstream detector localize the inharmonious region more easily. However, based on human perception, it may be arguable that the domain discrepancy between inharmonious region and background is enlarged after color transformation (column 3 \emph{v.s.} column 1). Here, we only provide some tentative interpretations as follows.
In the retouched images, we can observe that the false alarm (car in row 1) and the distractive objects (giraffe in row 2 and wood in row 3) are suppressed to some extent and the inharmonious region becomes more obtrusive. For example, in row 2, the distractive giraffe looks more similar to the middle giraffe while the inharmonious giraffe looks more different after transformation. In row 3, the distractive wood at the bottom right looks more harmonious with the wood on its left after transformation.

\section{Results on Four Sub-datasets}\label{sec:four_datasets}
Our used iHarmony4 \cite{cong2020dovenet} dataset consists of four sub-datasets: HCOCO,  HFlickr, HAdobe5K, HDay2night. We compare with the strongest baseline DIRL \cite{liang2021inharmonious} and report the results on four sub-datasets in Table~\ref{tab:sub_datasets}. Among them, HDay2night is a challenging sub-dataset with real composite images and has much fewer training images, leading to overall lower results. However, our method outperforms DIRL significantly on all four sub-datasets.

\section{Results for Multiple Inharmonious Regions} \label{sec:multiple_regions}

In this paper, we mainly focus on one inharmonious region. However, in real-world applications, there could be multiple disjoint inharmonious regions in a synthetic image. Here, we demonstrate the ability of our method to identify multiple disjoint inharmonious regions, starting from constructing usable dataset based on HCOCO. Specifically, in the sub-dataset HCOCO of iHarmony4 \cite{cong2020dovenet}, there exist some real images which have multiple paired synthetic images with different manipulated foregrounds. Based on HCOCO test set, we composite 19,482 synthetic images with multiple foregrounds for evaluation, in which the number of foregrounds ranges from 2 to 9. 
The results (AP, $F_1$, IoU) of DIRL and Ours are (73.62, 0.5079, 37.05) and (77.39, 0.5761, 44.03) respectively.
We also provide some visualization results in Figure~\ref{fig:multi_objects}, which show that our method can successfully localize multiple inharmonious regions.

\bibliographystyle{aaai22}
\bibliography{2.supplementary.bbl}